\documentclass{article}

\usepackage{PRIMEarxiv}

\usepackage[utf8]{inputenc} 
\usepackage[T1]{fontenc}    
\usepackage{hyperref}       
\usepackage{url}            
\usepackage{booktabs}       
\usepackage{amsfonts}       
\usepackage{nicefrac}       
\usepackage{microtype}      
\usepackage{lipsum}
\usepackage{fancyhdr}       
\usepackage{graphicx}       
\graphicspath{{media/}}     
\usepackage{booktabs}
\usepackage{multirow}
\usepackage{rotating}  
\usepackage{makecell}  
\usepackage{float}
\title{A Systematic Review on the Generative AI Applications in Human Medical Genomics
}

\author{
  Anton Changalidis, Yury Barbitoff, Yulia Nasykhova, Andrey Glotov \\ \\
  Dpt. of Genomic Medicine\\
  D.O. Ott Research Institute of Obstetrics, Gynaecology, and Reproductology\\
  St. Petersburg, Russia \\
  \texttt{anton@bioinf.me}, \texttt{barbitoff@bioinf.me}, \texttt{anglotov@mail.ru}
}

\begin{document}
\maketitle

\begin{abstract}
Although traditional statistical techniques and machine learning methods have contributed significantly to genetics and, in particular, inherited disease diagnosis, they often struggle with complex, high-dimensional data, a challenge now addressed by state-of-the-art deep learning models. Large language models (LLMs), based on transformer architectures, have excelled in tasks requiring contextual comprehension of unstructured medical data. This systematic review examines the role of LLMs in the genetic research and diagnostics of both rare and common diseases. Automated keyword-based search in PubMed, bioRxiv, medRxiv, and arXiv was conducted, targeting studies on LLM applications in diagnostics and education within genetics and removing irrelevant or outdated models. A total of $172$ studies were analyzed, highlighting applications in genomic variant identification, annotation, and interpretation, as well as medical imaging advancements through vision transformers. Key findings indicate that while transformer-based models significantly advance disease and risk stratification, variant interpretation, medical imaging analysis, and report generation, major challenges persist in integrating multimodal data (genomic sequences, imaging, and clinical records) into unified and clinically robust pipelines, facing limitations in generalizability and practical implementation in clinical settings. This review provides a comprehensive classification and assessment of the current capabilities and limitations of LLMs in transforming hereditary disease diagnostics and supporting genetic education, serving as a guide to navigate this rapidly evolving field.
\end{abstract}


\section{Introduction}
\subsection{Machine Learning, Deep Learning, and Language Models}

Machine learning (ML) has become a crucial tool in various fields, from healthcare to research, due to its ability to automate complex tasks and discover patterns in large datasets. Recent reviews highlight the growing impact of ML approaches in biomedical fields, including applications in diagnosing rare diseases and improving clinical outcomes \cite{healthcare10030541, Roman-Naranjo2023-uh}.

Traditional machine learning methods, such as decision trees and support vector machines, have been effective in solving well-defined problems where labeled data is abundant. However, these methods often struggle with high-dimensional data, complex relationships, and tasks that require context-dependent understanding, such as natural language processing (NLP) and genomics. One of the major challenges in traditional ML is handling large datasets with long-range dependencies -- where information far apart in the data sequence needs to be considered together to make accurate predictions. Additionally, it often relies on manual feature extraction and struggles with tasks that require a deeper context or understanding of relationships across the data. 

With the advent of deep learning (DL), many of these limitations were overcome. Deep learning, particularly with the use of neural networks, enables models to learn directly from raw data by automatically discovering useful patterns and representations. Convolutional Neural Networks (CNNs) excel at processing images \cite{Alzubaidi2021}, while Recurrent Neural Networks (RNNs) were initially used for sequential data like text \cite{ALSELWI2024102068}. However, RNNs also encountered difficulties with tasks that involved understanding relationships across long sequences of text due to their inherent sequential processing. This led to the development of transformer-based architectures, which revolutionized NLP and a range of other fields.

The introduction of transformer models in 2017 marked a significant breakthrough in deep learning \cite{vaswani2023attentionneed}. Unlike RNNs, transformers use an attention mechanism that allows the model to focus on different parts of the input data simultaneously, capturing long-range dependencies more effectively. This approach solves the problem of sequential processing and enables the model to understand complex relationships in data, very critical in healthcare and genomics. Transformers are particularly powerful in tasks that require context comprehension, such as text generation, translation, and named entity recognition. Their architecture consists of two main components: the encoder, which processes the input data (e.g., text or any other sequence, such as DNA), and the decoder, which generates the output (e.g., text). These terms refer to different stages of the model’s operation: encoding involves breaking down and analyzing input data to form a representation, while decoding reconstructs or predicts the next part of the sequence based on that representation.

BERT (Bidirectional Encoder Representations from Transformers) and GPT (Generative Pre-trained Transformer) are two of the most widely known transformer-based models, each tailored for different purposes. BERT is an encoder-only model, designed to understand text in both directions (left to right and right to left), which enables it to capture a more complete context for tasks like text classification and entity recognition. This bidirectional understanding allows the model to make more accurate predictions about the meaning of a word or phrase based on its surrounding context \cite{devlin2019bertpretrainingdeepbidirectional}. On the other hand, GPT is a decoder-only model that focuses on generating text, predicting each next word based on the preceding words in a unidirectional fashion. This makes GPT highly effective at tasks, such as text generation, translation, and summarization  \cite{radford2018improving}.

The ability of transformers to handle large datasets and maintain coherence over long sequences has led to the development of large language models (LLMs) – models with millions or billions of parameters \cite{brown2020languagemodelsfewshotlearners}. These models are capable of performing a variety of tasks by leveraging either full training on large datasets or fine-tuning with smaller, task-specific datasets. Fine-tuning allows the model to adapt to new tasks with minimal additional data, making few-shot or one-shot learning techniques possible: in few-shot learning, the model requires only a few labeled examples to perform well, while in one-shot learning, it can generalize from just a single example. This adaptability enables LLMs to be highly efficient across a range of applications, including research, healthcare, and education, without the need for retraining from scratch \cite{Du2024,genes15040421,aronson2024preparingintegrategenerativepretrained, Ueda2024, Laye2024-qg}. 

Vision Transformers (ViTs) have further extended this approach beyond text, applying the transformer architecture to image processing tasks \cite{dosovitskiy2021imageworth16x16words}. By treating image patches like words in a sentence, ViTs can capture dependencies across different parts of an image, making them highly effective in tasks like image classification and segmentation. The versatility of transformers across multiple domains demonstrates their power and adaptability, making them integral to modern AI applications. The versatility of transformers across multiple domains demonstrates their power and adaptability, making them integral to modern AI applications.

Generative Adversarial Networks (GANs) complement this landscape as specialized models for data generation, enabling the synthesis of highly realistic images, biomedical data, and even artificial genetic sequences through adversarial training.

Meanwhile, foundation models are trained on vast and diverse datasets and subsequently adapted (fine-tuned) to a wide variety of downstream tasks with minimal task-specific data. They shape the backbone of modern AI, providing general-purpose representations that can be adapted to a variety of specialized tasks. These models excel in transferring learned knowledge to new domains, accelerating advances in research, healthcare, and genomics.

\subsection{Inherited disease diagnostic pipeline}
\label{intro_dis}

Laboratory diagnostics of inherited disease has two major goals that are inherently interconnected: (i) establishing the correct diagnosis of the disease or syndrome affecting the patient; and (ii) finding the exact genetic cause of the condition. The former goal involves the analysis of the patient’s phenotype and healthcare data, and, in certain cases, can not be achieved without access to the genetic data. For the latter goal of finding the genetic cause of the disease, data (usually, coming from next-generation sequencing (NGS)) are processed in order to identify the causal genetic variation (a single-nucleotide substitution, a short insertion or deletion, or a more complex alteration of a genome sequence) (reviewed in \cite{10.1093/bib/bbad508}). Similarly, the search for the genetic cause is greatly aided by the patient’s phenotypic data and/or candidate syndrome information. For complex diseases affected by multiple genetic variants, the data processing step is typically focused on evaluating the overall genetic risk of a condition rather than on the identification of individual causal variants \cite{Wand2021-dn}.

A typical diagnostic workflow in medical genetics can be divided into three key stages -- pre-analytical, analytical, and post-analytical -- each of which involves specific tasks and methods. Pre-analytical stage focuses on patient preparation and data organization, including gathering patient history and clinical information, assessing genetic predispositions, as well as biological sample collection and preparation. Additionally, it can include a literature review and knowledge aggregation for the specific field or study.

The analytical stage is the core diagnostic phase, where data is processed and interpreted. For NGS data, this stage involves calling genomic variants using bioinformatics tools, followed by the annotation of the discovered variants. The latter step frequently requires manual curation and heavily relies on predefined external datasets and resources. Moreover, in addition to genetic data, information from other modalities, such as Magnetic Resonance Imaging (MRI) and computed tomography (CT) scans, histology, and patient photographs, can provide valuable complementary insights into genetic predispositions.

Post-analytical stage involves integrating diagnostic results into clinical decisions: subtyping diseases and clustering patients, as well as results aggregation, decision making, and report generation.

While recent reviews have explored the potential of artificial intelligence and, more specifically, transformer models in healthcare and genomics, many have limitations in scope or model specificity. For example, some reviews focus solely on the applications of ChatGPT without a systematic analysis \cite{genes15040421, wang2024bioinformaticsbiomedicalinformaticschatgpt, Jeyaraman2023-dg}, making them outdated or too narrowly focused. Broader reviews, such as those on LLMs in general healthcare applications, lack specific emphasis on genetic diagnostics \cite{Bedi2025-tp, 10.1093/bib/bbae156}. Some reviews are limited to a specific disease, such as dementia \cite{moya2024addressinggapsearlydementia}, oncology \cite{Webster2023-of, Mudrik2024.08.08.24311699}, schizophrenia \cite{jpm14070744} and often does not have a clear emphasis on transformer-based models\cite{Venkatapathappa2024-kq, dai2024identifyinghealthrisksfamily}, moving a the scope of insights away from LLMs for genetic data analysis. 

While the closest topical review broadly covers AI in clinical genetics, it focuses on conventional DL methods and lacks depth on LLMs and transformers \cite{Duong2025-xi}. It is also not systematic or comprehensive, limiting its value as a foundational reference. This systematic review focuses specifically on the application of transformer models and generative AI in the research and diagnosis of hereditary diseases in recent years. To provide a comprehensive perspective, we reviewed models from four key sources: PubMed, bioRxiv, medRxiv, and arXiv, thus including both peer-reviewed studies and the latest preprint models. Since many state-of-the-art models are initially released as open-source in preprint repositories, this approach ensured we did not overlook recent developments. The growing need for efficient data processing and analysis in these domains highlights the potential of LLMs to revolutionize our understanding of genetic data, improve diagnoses, and predict disease outcomes. By exploring the use of LLMs in pre-analytical, analytical, and post-analytical stages, this review aims to provide systematic insights into how these models are transforming diagnostics, automating clinical processes, and supporting personalized medicine. A dedicated section will also assess the performance of GPT-like models in clinical and research settings, examining both effective and problematic practices.

\section{Methods}

\begin{figure}
\centering
\includegraphics[width=0.55\textwidth]{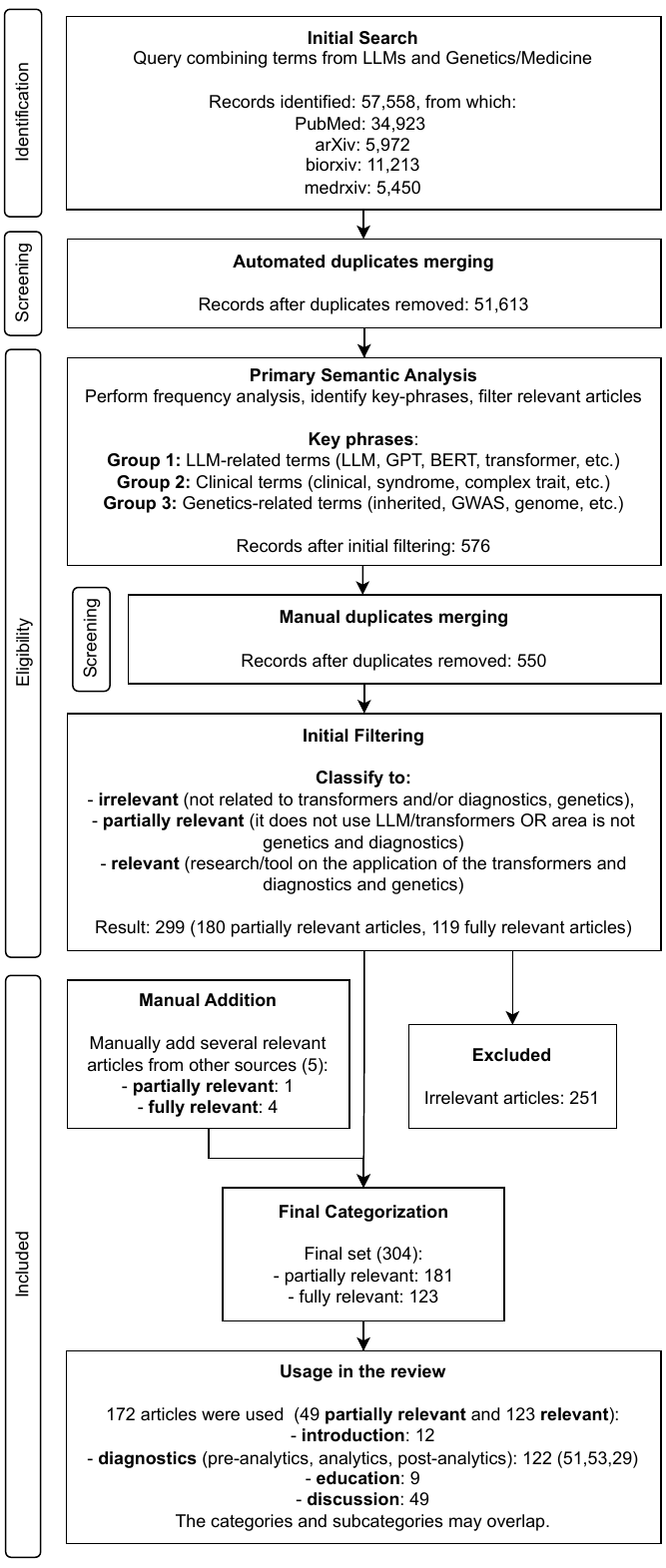}
\caption{Pipeline of search strategy and filtering of the articles.\label{fig1}}
\end{figure}

To comprehensively analyze the use of transformer-based models in genetics and hereditary diseases, a systematic review approach was developed according to the latest PRISMA 2020 guidelines for reporting systematic reviews \cite{Pagen71}, ensuring thorough and transparent coverage of relevant studies. The search strategy was carefully constructed with selected terms relevant to transformer-based models and genetics, and all records were evaluated through a consultative process by two researchers, allowing for in-depth discussions on ambiguous cases, promoting a balanced selection, and reducing potential bias. The full search process is visualized in Figure \ref{fig1}.

\subsection{Search strategy}

To systematically review the use of LLMs in genetics and hereditary diseases, an initial broad search for relevant articles in English was conducted across multiple major scientific databases. A custom Python script was developed to automate the collection of articles from PubMed, bioRxiv, medRxiv, and arXiv (see \textit{Data Availability} for access to the code repository). The search criteria focused on articles from 2023, 2024, and the beginning of 2025 (January) to ensure the inclusion of the most up-to-date research in this rapidly evolving field (the dataset was downloaded on 31-01-2025). Articles from medRxiv and bioRxiv were accessed through the API available at \url{https://api.biorxiv.org/} (accessed on 31-01-2025), while arXiv data was retrieved using the Python wrapper \url{https://github.com/lukasschwab/arxiv.py} for the arXiv API (accessed on 31-01-2025). PubMed articles were accessed via the Biopython package for the PubMed API \cite{cock2009biopython}, available at \url{https://biopython.org/docs/1.76/api/Bio.Entrez.html} (accessed on 31-01-2025). This process yielded an initial dataset of 57,558 articles, forming the basis for further analysis.

The query terms were divided into two groups: one related to genetics and medicine, and the other related to transformer models and LLMs. Relevant articles were required to contain at least one term from each list in their title and/or abstract:

\begin{itemize}
    \item genomic, genetic, inherited, hereditary, heredity, inheritance, heritability, disease subtype, NGS, next-generation sequencing, next generation sequencing, genome sequencing, phenotype description, variant interpretation, complex trait, medicine, medical, diagnosis, diagnostic, clinical, clinical decision, syndrome.
    \item LLM, large language model, NLP, natural language processing, GPT, chatGPT, transformer, BERT, Bidirectional Encoder Representation, RAG, retrieval-augmented generation, retrieval augmented generation, generative AI, AI assistant, prompt, chatbot, prompt engineering, attention mechanism, chain-of-thought, chain of thought.
\end{itemize}

\subsection{Inclusion and Exclusion Criteria}

\begin{figure}[h]
\includegraphics[width=\textwidth]{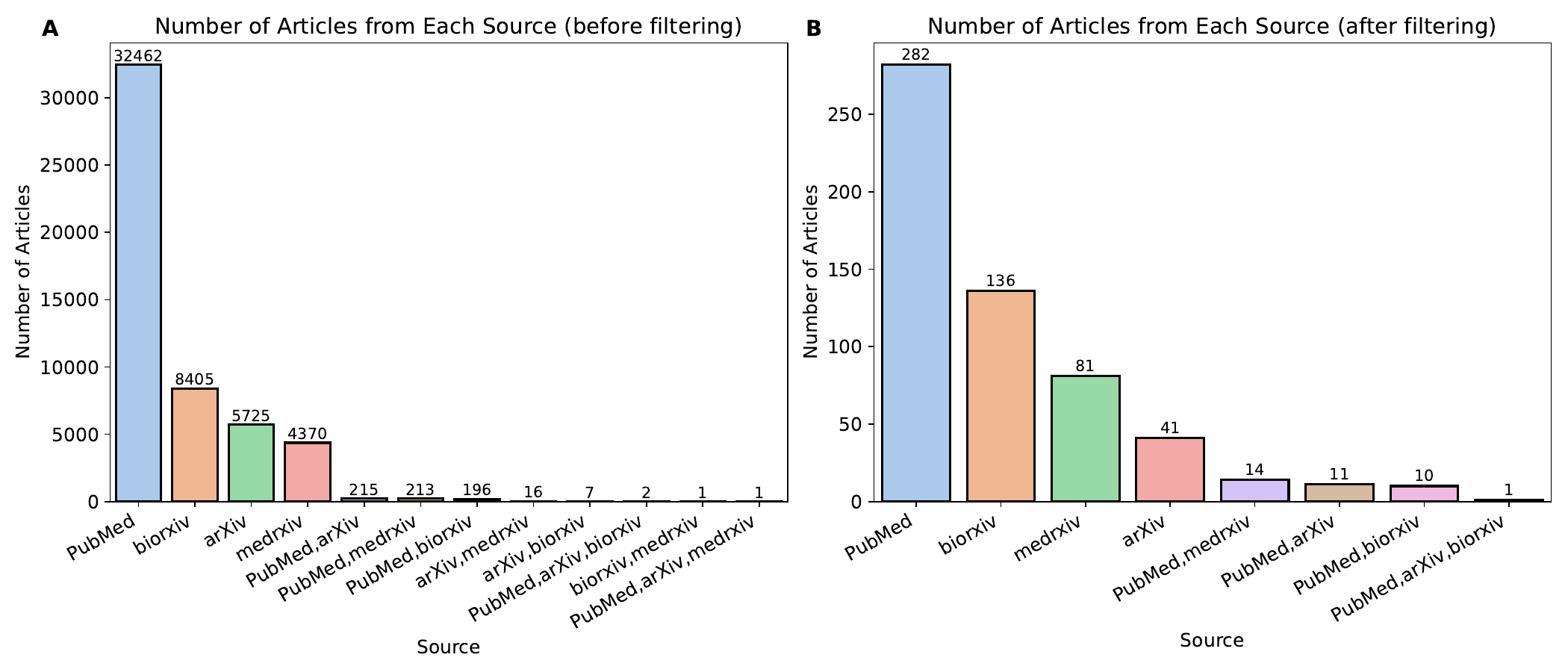}
\caption{Distribution of articles by source: (A) after automatic deduplication and merging (51,613 records in total); (B) after additional automated filtering for relevance to clinical diagnostics (576 records in total).
\label{fig2}}
\end{figure}

After retrieving articles, several steps of filtering and exclusion were conducted. The first step in data processing involved automatically removing duplicate entries and cleaning the data, reducing the dataset to $51,613$ articles. This was done using text processing algorithms to detect similarities in titles and abstracts. Figure \ref{fig2}A illustrates the contribution of each database to the final dataset, with a substantial number of preprints included. Although preprints offer access to the latest research, they lack peer review and may contain unverified results, requiring careful analysis.

A primary semantic analysis was performed to assess the relevance of each article to the research objectives. To identify domain-specific vocabulary, TF-IDF (Term Frequency–Inverse Document Frequency) scores were calculated for all words and phrases found in article titles and abstracts. This helped highlight key terms related to genetics, hereditary diseases, and large language models (LLMs), which are visualized in Supplementary Figures \ref{SF1}–\ref{SF2}. The identified phrases were grouped into three semantic categories:

\begin{itemize}
    \item LLM-related terms: LLM, large language model, NLP, natural language processing, GPT, chatGPT, transformer, BERT, Bidirectional Encoder Representation, RAG, augmented generation, generative AI, AI assistant, prompt engineering, chatbot, prompt engineering, attention mechanism, chain-of-thought, chain of thought.
    \item Clinical terms: electronic health record, ehr, clinical, case report, cds, intensive care unit, medical, syndrome, phenotype, complex trait.
    \item Genetics-related terms: inherit, heredit, heritability, gwas, genome-wide, genome wide, association stud, snp, single nucleotide, genetic, variant interpretation, genomic varia, human gen, NGS, generation sequencing.
\end{itemize}

To ensure coverage of morphologically derived forms (e.g., “inherited”, “genomics”, “associations”), the terms above were defined using stemmed substrings and matched via regular expressions. Filtering required that each article contain at least one match from each of the three categories.

To avoid false positives caused by accidental substring matches in unrelated words (e.g., “cove\textbf{rag}” or “encou\textbf{rag}” falsely matching “rag”), an empirically derived exclusion list was applied. This list was constructed by manually reviewing articles irrelevant to the study focus and identifying recurring misleading terms. This list included the following terms or common letter combinations: \textit{tragic, fragment, coverag, encourag, ungs, angs, ongs, ings, eragrostis, smallmouth, fragile, angptl, intragenic, fragment, hallmark, uvrag, leverag, storag, averag, coverag, encourag, forage, liraglutid}. This filtering strategy significantly improved the precision of semantic classification by excluding structurally similar but contextually irrelevant terms.

Additionally, a manual verification step was conducted to identify and remove duplicate entries that were not detected automatically. In several cases, articles had slightly different titles or abstracts but were authored by the same group and described the same study. Based on this content-level similarity and author overlap, duplicates were removed, reducing the dataset from 571 to 550 articles for subsequent analysis. The complete list of included articles is provided in Supplementary Table 1. As previously noted, this step, as well as all subsequent ones, were conducted jointly by two researchers, allowing for careful discussion of ambiguous cases and minimizing potential bias.

After deduplication, articles were manually divided into three classes, based on their relevance to the topic:

\begin{itemize}
    \item Irrelevant articles (not included in the review): 251 articles were excluded as they did not align with the study's focus on transformer-based models in genetics or were outdated.
    \item Partially relevant articles: 180 articles discussed either the application of language models but were not directly related to the study's focus, or the usage of outdated models (e.g., not transformer-based) for the relevant tasks.
    \item Fully relevant articles: 119 articles directly aligned with the research goals, utilizing modern transformer architectures in diagnostics and genetics.

\end{itemize}

In addition to automated filtering, five manually selected articles of partial (1) and high (4) relevance were included in the final dataset, bringing the total to 304 articles (Supplementary Table 2): 181 partially relevant articles and 123 fully relevant articles. Additional articles were sourced through references from the initially selected studies, as well as through further targeted filtering and searches across the originally extracted dataset. 

At this stage, a thorough investigation of the selected articles was conducted. All highly relevant articles, as well as some of the partially relevant ones, were included in the analysis. From the latter category, only those entries were chosen that provided good examples of deep learning methods used in diagnostics, even if not specifically focused on LLMs or transformers. During the review process, each article was assessed based on its relevance to specific sections outlined in \textit{1.2} of 
 \textit{Introduction} (pre-analytical, analytical, and post-analytical stages), as well as educational uses of LLMs in genetics. Following classification, the review was organized into two main sections: (1) analytical applications of LLMs in genetics, divided into three stages, and (2) educational applications, focusing on how LLMs can support the learning process for professionals in genetics. Additionally, insights into best and worst practices of transformer usage are outlined in the discussion section. Since these areas encompass a broad list of tasks, they have been divided into specific applications, and itemized (see the relevant sections, Table \ref{tab_bar} and Figure \ref{fig3}).

\begin{table}[ht]
\centering
\begin{tabular}{l llr}
\toprule
& \textbf{Article section} & \textbf{Research/application area} & \textbf{Articles count} \\
\midrule

\multirow{1}{*}{} 
    & Introduction  
        & review                                               & 12 \\
\midrule

\multirow{8}{*}{\rotatebox{90}{\parbox{2.2cm}{\centering Diagnostics}}}
    & \multirow{2}{*}{Pre-Analytical stage} 
        & knowledge navigation \& literature review            & 40 \\
    &   & risk stratification                                  & 11 \\
    \cmidrule{2-4}
    & \multirow{3}{*}{Analytical stage}     
        & medical imaging analysis                             & 24 \\
    &   & analysis of variant effects                          & 23 \\
    &   & clinical variant interpretation                      & 9  \\
    \cmidrule{2-4}
    & \multirow{3}{*}{Post-Analytical stage}
        & patient clustering \& subtyping                 & 7  \\
    &   & data \& results aggregation                          & 8  \\
    &   & clinical report generation \& decision support       & 15 \\
\midrule

\multirow{1}{*}{} 
    & Education       
        & education                                            & 9  \\
\midrule

\multirow{1}{*}{} 
    & Discussion      
        & specificity of LLM usage, other medical topics       & 49 \\
\bottomrule
\end{tabular}
\caption{Distribution of article usage across sections and application areas. Articles may appear in multiple categories.}
\label{tab_bar}
\end{table}

In total, of the $304$ selected studies, $172$ were used ($123$ relevant and $49$ partially relevant). 
Among these, 122 focused on diagnostics, 9 on education, and 49 were used as examples of practices discussed in the review 
(with some articles used in multiple sections). 
Furthermore, 12 articles were incorporated in the introduction as examples of existing systematic research with a similar topic (see Table~\ref{tab_bar}).

\subsection{Risk of Bias}
 A large proportion of the selected articles came from preprint databases, such as arXiv, bioRxiv, and medRxiv, meaning they had not yet undergone peer review. This could introduce some bias, as these studies have not been validated by the scientific community. However, given the fast-paced nature of LLM development, many of the most cutting-edge techniques are being developed faster than the peer-review process allows. Consequently, it was deemed essential to include such articles to capture the most current advancements.

Additionally, while this review focuses on the application of LLMs in the specific domain of genetics and hereditary diseases, there may be general-purpose models or methods from broader AI fields that were not included in this focused analysis. These models could still provide valuable insights or advancements applicable to this domain, although they fall outside the scope of this particular review.

\section{Results}

\subsection{Applications of LLMs in inherited disease diagnostics}

\begin{figure}
\centering
\includegraphics[width=\textwidth]{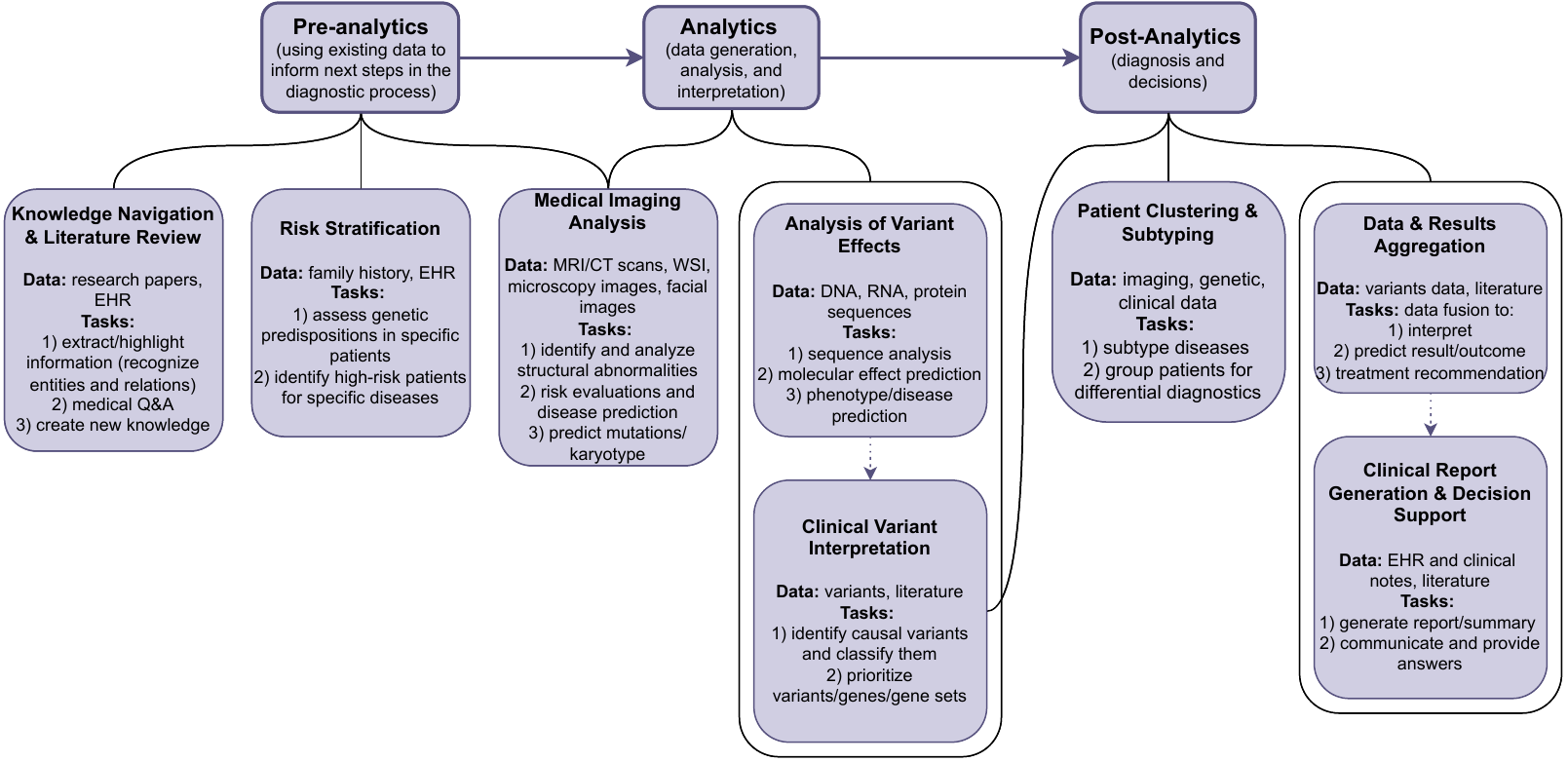}
\caption{A diagram showing the applications of LLMs in the research and diagnosis of human genetic diseases.\label{fig3}}
\end{figure} 

We now examine each phase of the diagnostic pipeline, as introduced in Section~\ref{intro_dis} and illustrated in Figure~\ref{fig3} and Table~\ref{tab_bar}. In this systematic review, a total of 51 articles focused on the pre-analytical stage. Within this phase, 39 articles addressed knowledge navigation and literature review (extraction of structured information, creation of new one, using research texts and electronic health records (EHRs)), and 12 examined risk stratification (use genetic or phenotypic data to assess genetic predispositions or identify high-risk patient groups). The analytical stage received the most attention, with 53 studies in total. These were distributed across three major tasks: 24 articles for medical imaging analysis (detecting pathological features in visual data), 23 articles for analysis of variant effects (sequence analysis, predicting molecular or phenotypic consequences of variants), and 9 articles for clinical variant interpretation (identifying, classifying and prioritizing causal variants using multimodal inputs). The post-analytical phase was the focus of 29 articles. These included 7 studies on patient clustering and subtyping (based on analytical stage data), 8 on data and results aggregation (data interpretation for prediction and recommendation), and 15 on clinical report generation and decision support (providing explainable outputs to physicians or patients). This distribution highlights the increasing integration of LLMs across all stages of the diagnostic workflow, with an emphasis on the pre-analytical and analytical phases. However, they remain supportive tools in decision-making, with human expertise still essential.

\subsubsection{Pre-analytical stage: enhancing next steps through existing data}

The pre-analytical phase includes all preparatory steps undertaken prior to genetic analysis, focusing on the extraction and organization of relevant information. LLMs and transformer-based systems are used here to retrieve general biomedical knowledge from the literature, extract patient-specific details from EHRs, and contextualize individual and familial risk factors. These tools help identify high-risk individuals, assess predispositions, and frame the clinical context necessary for downstream analyses. Moreover, they support clinicians in structuring complex case histories and in aligning data with the goals of subsequent analytical steps.

\paragraph{Knowledge Navigation and Literature Review.}

In this section, we describe how LLMs and transformer-based models are applied to explore, retrieve, and synthesize biomedical knowledge from literature, structured databases, and EHRs. Key tasks include extracting entities and relationships (e.g., genes, phenotypes, variants), answering complex clinical questions, and generating or enriching biomedical knowledge graphs. These models aid clinicians and researchers in hypothesis formation, evidence retrieval, and data contextualization. While encoder-based models dominate extraction tasks, recent studies have also explored generative LLMs for question answering and hypothesis generation, highlighting both their potential and current methodological trade-offs.

The first area encompasses diverse efforts to extract structured knowledge from biomedical literature and clinical texts, particularly focusing on gene, variant, disease, and phenotype recognition. A large number of studies address Human Phenotype Ontology (HPO) \cite{Gargano2024-uh} extraction and normalization using both encoder- and decoder-based models \cite{YANG2024100887, albayrak2025100409, murphy2024.06.10.24308475, 10.1093/database/baae103, 10.3389/fdgth.2025.1495040, Weissenbacher2024-gm}. Notably, several works fine-tune BERT-based architectures for named entity recognition (NER) and relation extraction (RE), achieving high scores in tasks such as identifying gene-disease or phenotype-variant relationships \cite{huang2024-vl, 10.3389/fnins.2023.1266771, lee2023-pz}. More advanced pipelines incorporate active learning for dataset creation (DUVEL (Detection of Unique Variant Ensembles in Literature)), expanding coverage to complex multi-entity relations of genes and variants \cite{10.1093/database/baae039}. In parallel, other tools (GPAD (Gene-Phenotype Association Discovery), RelCurator) specifically target gene-disease associations and neurodegenerative disorder phenotypes, respectively, offering curated resources and interpretable extraction frameworks \cite{rahit2024, lee2023-pz}. Some works develop hybrid methods combining retrievers with LLMs, achieving enhanced accuracy in HPO normalization by embedding-based matching \cite{10.3389/fdgth.2025.1495040}. However, there is a rising trend of employing decoder-based LLMs (e.g., GPT-3.5/4, PhenoGPT, GP-GPT) for entity-level tasks, which, despite promising results in some studies, may be suboptimal for structured extraction due to architectural mismatches \cite{murphy2024.06.10.24308475, lyu2024gpgptlargelanguagemodel, YANG2024100887}. This trend invites further research, as will be described in the Discussion. Additionally, some research shows that open-source models work better than ChatGPT in the biology domain. These models, unlike ChatGPT, have an additional benefit of possible fine-tuning for downstream tasks \cite{Khoury2024.09.13.612922}. Importantly, several contributions highlight the importance of cross-institutional validation \cite{sushil2024-jf} and model ensemble strategies, such as the Two-Stage LLM framework for adaptive entity matching across heterogeneous datasets \cite{xue2024-qa}. Together, these works illustrate both the power and evolving challenges of applying transformer models to biomedical entity and relation extraction in the pre-analytical phase.

The second area of LLMs applications supports biomedical and clinical question answering, from general literature-based fact retrieval to fine-grained evidence extraction for variant interpretation. Tools, such as PubTator 3.0 combine large-scale pre-computed entity annotations with LLM-augmented APIs, significantly improving literature navigation and answer factuality \cite{10.1093/nar/gkae235}. Several studies develop or fine-tune domain-specific Q\&A systems, such as BioMedLM \cite{bolton2024biomedlm27bparameterlanguage}, a GPT-style model trained on PubMed data for medical tasks. Others, like ClinVar-BERT and AutoPM3, optimize question answering specifically for variant prioritization and interpretation, using BioBERT-style models and retrieval-augmented generation (RAG) to extract information from publications \cite{Li2024-ul, Li2024.10.29.621006}. As shall be discussed later in the review, the same approaches are used not only for retrieving general information about a given variant but also to interpret its effect in a specific patient. In complex disease contexts, models such as BRLM (BioBERT vectorized input for ResNet classification Language Model) \cite{WANG2024843} and Deep-GenMut \cite{ELSAMAHY2024e32279} integrate BioBERT with deep classifiers for identifying mutation classes and gene-phenotype associations. Meanwhile, GeneGPT augments Codex \cite{chen2021evaluatinglargelanguagemodels} with live NCBI API access, outperforming biomedical LLMs on genomic tasks via tool-enhanced prompting \cite{10.1093/bioinformatics/btae075}. Q\&A systems are also benchmarked on real-world and layperson input: one evaluation compared 11 LLMs on recognizing genetic disorders from professional vs. self-reported descriptions, finding GPT-4 the most robust but highlighting performance drops with less formal queries \cite{FLAHARTY20241819}. Several studies test GPT-4 in clinical-grade workflows (e.g., for evidence grading, variant classification), warning that nondeterminism and model drift must be addressed before deployment in diagnostic settings \cite{aronson2024preparingintegrategenerativepretrained}. Finally, FindZebra tool extends NLP-enhanced search over rare disease case reports, enriching diagnostic support through patient-specific semantic queries \cite{Lievin2023-pw}, while Med-PaLM 2 \cite{singhal2023expertlevelmedicalquestionanswering} shows promise in enabling genetic discovery from free-text gene lists in preclinical research \cite{Tu2023.11.09.566468}. Together, these works show that biomedical Q\&A with LLMs is evolving toward increasingly task-specific, tool-integrated, and clinically aligned applications.

The last area captures efforts where transformer-based models are used not merely for extraction, but for generating new biomedical knowledge, hypotheses, or structured representations. A common thread is the construction and use of graph-based models, including condition-specific knowledge graphs, such as CPMKG (Condition-based Precision Medicine Knowledge Graph) for drug-disease-phenotype reasoning \cite{10.1093/database/baae102}, hierarchical or hypergraph architectures for biomarker and therapeutic gene prediction \cite{Zhang2023.01.30.23285175, 10.1093/bib/bbaf019}, and transformer-enhanced model DGP-PGTN (Disease–Gene association Prediction model with Parallel Graph Transformer Network) for disease-gene association prediction \cite{10.1093/bib/bbad118}. Several works applied contrastive or multimodal learning, such as COMICAL (contrastive multi-omics association learning) for linking genetic markers with brain imaging phenotypes \cite{Machado_Reyes2024.11.02.24316653}, and LitGene, which uses contrastive learning to derive semantically grounded gene representations by aligning textual descriptions with Gene Ontology labels \cite{Jararweh2024.08.07.606674}. Others mined vast literature corpora to derive latent gene-disease links, and fine-tuned like PathoBERT, enabling vector-based hypothesis generation: predict disease-gene associations that are not cited in the training data \cite{10.1093/bioinformatics/btae185}. LLMs also play a growing role in augmenting biocuration: from AutoMAxO’s (MAxO - Medical Action Ontology) semi-automated annotation for rare diseases \cite{Niyonkuru2024.08.22.24310814}, to using LLMs for dynamic disease discovery and providing genetic insights \cite{genes16010029, chang2024geneassociateddiseasediscoverypowered}. Finally, works like SCREENER (Streamlined CollaboRativE lEarning of NEr and Re) \cite{Park2023-ec} and the SemNet-based dementia study \cite{ijms252413450} illustrate how attention-based models and unsupervised graph aggregation can support novel relationship discovery across large biomedical corpora. Collectively, these methods demonstrate that transformer models are being increasingly adapted not only to interpret biomedical data but to construct new layers of structured, testable knowledge, a pivotal shift in computational genetics.

\paragraph{Risk Stratification.}

In this subsection, we focus on using transformer-based models and LLMs to evaluate disease risk through clinical notes, family history, and multimodal health data. Two core tasks are addressed: assessing individual genetic predisposition from phenotypic or textual input, and identifying high-risk patients using predictive modeling across large populations. Approaches range from using LLMs to prioritize candidate genes or suggest differential diagnoses, to modeling longitudinal health records, imaging, and family relations. 

Several studies have explored the feasibility of using LLMs to assess inherited disease risk by interpreting clinical notes, patient histories, and phenotypic descriptions. These works typically apply generative LLMs such as GPT-3.5, GPT-4, and Gemini: e.g., suggesting candidate diagnoses in autoinflammatory and neurogenetic disorders, or predicting cancer predisposition genes from textual EHR summaries \cite{PILLAI2023100213, genes16010029, Sultan2023-eu}. Some studies show that LLMs can prioritize causative genes directly from free-text phenotypes, matching traditional ontology-based methods without relying on structured databases \cite{Kafkas2023.11.16.23298615}. While generative models perform well on curated inputs, they still face issues like hallucinations and prompt sensitivity, highlighting both their promise and current limitations in clinical risk assessment.

Models with attention mechanisms can also identify patients at elevated risk for specific diseases by modeling diverse inputs such as EHRs, imaging, genetics, and clinical notes. Some models focus on predicting long-term disease trajectories: heart failure risk in congenital heart disease patients using masked self-attention over medical events \cite{MOROZ2024105384}, or survival risk in breast cancer through multimodal fusion of pathology images, genetic data, and clinical records \cite{mondol2024mmsurvnetdeeplearningbasedsurvival}. Others develop multi-task learning systems that combine segmentation and treatment response prediction, integrating transformer modules for spatial reasoning in medical imaging \cite{LIU2024108503}. Graph-based approaches have also been applied to incorporate predicted family relations into risk modeling, enhancing disease prediction from EHR data \cite{Huang2024.03.12.24304163}. Finally, benchmarking studies assessing public LLMs on complex breast cancer treatment decisions show that newer models like GPT-4 outperform earlier ones, though current limitations in reliability and clinical readiness remain evident \cite{Griewing2024}. Together, these efforts demonstrate how transformer-driven systems can augment risk stratification by integrating temporal, relational, and multimodal biomedical data.

\subsubsection{Analytical stage: data analysis, and interpretation}

The analytical phase focuses on the core processes where raw biomedical data is transformed into clinically actionable insights using transformers. This phase is divided into three interconnected domains. The first is medical imaging analysis, where models interpret various visual representations to detect abnormalities, predict syndromes, and assess risks. The second is an analysis of variant effects, where models are applied to biological sequences to predict the effects and outcomes, followed by the third area, clinical variant interpretation: identify, classify, and prioritize variants, genes, or gene sets. Together, these domains represent the central analytical functions of LLM-driven genomics, bridging data-intensive tasks with downstream clinical decision making.

\paragraph{Medical Imaging Analysis.}

Transformer-based models are increasingly used in medical imaging to support genetic diagnostics, linking visual features to genomic data, identifying phenotypic patterns of genetic disorders, and even predicting mutations directly from histology or cytogenetics. This section covers three key directions: structural abnormality detection, risk evaluation through multimodal imaging, and direct genetic prediction from medical images.

Although structural abnormalities detected via medical imaging are not always direct markers of genetic disease, they can reveal phenotypic manifestations of underlying mutations or syndromes with genetic components. Recent studies used multimodal frameworks to link imaging with genetic data: MGI (Multimodal contrastive learning of Genomic and medical Imaging) aligned gene expression and MRI via contrastive learning for tumor segmentation \cite{zhou2024mgimultimodalcontrastivepretraining}, while BioFusionNet combined histology, genomic, and clinical features via cross-attention between different data types to improve cancer risk prediction \cite{Mondol_2024}. In the context of Alzheimer's disease, a common neurodegenerative disease with heritable risk factors \cite{A_Armstrong2019-wg}, researchers presented a feature-fusion ViT model adapted for low-resource environments, demonstrating that modified transformers can deliver interpretable and accurate MRI-based classification even without genetic input—while also emphasizing the future potential of integrating clinical and genomic data \cite{diagnostics14212363}. In a clinically grounded example, CKD-TransBTS (Clinical Knowledge-Driven TRANSformer for Brain Tumor Segmentation) is a hybrid CNN-transformer architecture using modality-correlated cross-attention, guided by radiological expertise to segment brain tumors more effectively \cite{Lin2023-yd}. Crucially, Conze et al. (2024) extended this paradigm to autosomal-dominant polycystic kidney disease, a monogenic disorder, demonstrating that ViT-based architecture showed strong performance in identifying and outlining cyst-affected kidneys in MRI scans, even when the cysts varied widely in size and location \cite{CONZE2024102349}. 

Another prominent direction is assessing genetic risk by analyzing visual patterns and integrating multiple data types. One great example is the use of facial image analysis to assist in diagnosing rare genetic conditions. DeepGestalt identified syndromic features in hundreds of conditions. However, it used a convolutional neural network architecture, not transformers, demonstrating the potential of AI in phenotype-based diagnostics \cite{Gurovich2019}. GestaltMML (MML - multimodal machine learning) extended this by combining facial photos, clinical notes, and metadata using Transformers, improving accuracy and aiding genomic interpretation \cite{wu2024gestaltmmlenhancingraregenetic}. Patel et al. (2024) used GAN-inversion to show that facial expressions like smiling can bias diagnostic accuracy for syndromes such as Williams and Angelman \cite{Patel2024-jf}. Beyond facial features, models like CroMAM (CROss-Magnification Attention feature fusion Model) employed histopathology images of gliomas to predict gene mutation status and patient survival by combining multi-magnification slide data through Swin Transformers (specific transformer architecture designed for computer vision tasks \cite{liu2021swintransformerhierarchicalvision}) and cross-level attention to extract local and global features from patches \cite{10605027}. These studies demonstrate how transformer-inspired architectures help link phenotype, genotype, and clinical context, offering generalizable methods for precision diagnostics across diverse fields in medicine.

Both in oncology and beyond, ViTs are used to identify genetic mutations, tumor molecular subtypes, and chromosomal abnormalities directly from medical images. Several studies apply attention mechanisms and Vision Transformers to predict gene mutations related to cancer using whole slide images \cite{Singh2024-ut, Sun2024, GUO2023102189}, while others (e.g. BPGT (Biologicalknowledge enhanced PathGenomic multi-label Transformer) \cite{huang2024predictinggeneticmutationslide} and PromptBio \cite{zhang2024promptingslideimagebased}) additionally integrate whole slide imaging (WSI) data with biomedical text, gene and variants descriptions, LLM-generated prompts to further enhance these predictions. Genetic InfoMax bridges human brain 3D MRI data and human genetic data to improve the performance of existing GWAS methods \cite{xie2023geneticinfomaxexploringmutual}. In another related area of research, ChromTR \cite{Xia2024} and Tokensome \cite{zhang2024tokensomegeneticvisionlanguagegpt} apply transformer models to microscopic metaphase images for karyotyping of the patient. Additionally, other studies \cite{Akram2024-zn, 10083175} focus on tumor subtyping and aneuploidy detection from histological imaging at the cellular level. Together, these works show how transformers enable direct genetic prediction from visual data, supporting more precise and interpretable diagnostics.

\paragraph{Analysis of Variant Effects.}

Transformer-based models are enabling a deeper understanding of how genetic variants influence biological function and disease. In this section, we explore three key directions: using language models to extract structure from raw DNA sequences; predicting the functional consequences of variants on proteins, regulation, and drug response; and modeling complex genotype–phenotype relationships for disease risk and trait prediction. These advances highlight the growing role of transformers in bridging molecular data and clinical insights.

Transformer-based foundation models are rapidly advancing sequence-level understanding in genomics by learning from raw DNA or methylation data at scale. MethylGPT demonstrates this by modeling human methylation profiles using a transformer to capture both local context and long-range chromatin features, enabling downstream predictions like age and disease risk \cite{Ying2024-vf}. Similarly, GENA-LM \cite{Fishman2023.06.12.544594} and Nucleotide Transformer \cite{Dalla-Torre2025} are long-context DNA foundation models, which offer robust representations that can be fine-tuned for a wide range of downstream tasks (e.g., prediction of splice sites, epigenetic marks, enhancer sequence, promoter sequence, enhancer activity, chromatin profile, and others). Such foundation models are especially useful in low-data or low-resource settings, where fine-tuning benefits from pre-trained sequence-level general knowledge. Complementing these, another study critically examines how DNA language models handle tokenization and sequence context. It shows that models trained on overlapping k-mers tend to memorize short token patterns rather than capture broader DNA context, making them well-suited for tasks focused on local features but less effective for those requiring long-range understanding \cite{Sanabria2024}. These models lay the foundation for complex tasks in genomics by extracting deep structure from sequences alone, following advances in natural language understanding.

Going from just analyzing DNA sequences, transformers are used, based on this analysis, to predict the functional consequences of genetic variants, particularly on protein structure, interaction, regulation, and drug response. AlphaMissense achieves state-of-the-art performance in classifying missense variant pathogenicity by combining structural data from AlphaFold \cite{Jumper2021} with unsupervised protein language modeling \cite{doi:10.1126/science.adg7492}. MIPPI (Mutation Impact on Protein–Protein Interaction) adds interpretability by modeling the effects of mutations on protein-protein interactions through attention patterns learned from sequence alone \cite{Liu2023-vl}. INTERACT applies transformer-based models to predict brain cell type-specific DNA methylation changes from variants, helping to fine-map causal variants in neuropsychiatric disorders \cite{Zhou2024-hq}. Similarly, EN-TEx demonstrates that a transformer can predict allele-specific regulatory activity across multiple tissues using only local sequence context, aiding the transfer of eQTL (expression quantitative trait loci) signals to less accessible tissues \cite{Rozowsky2023-hd}. Other works extend this framework to diverse clinically relevant tasks. For example, Emden integrates graph and transformer features to predict how missense mutations alter drug response \cite{LIU2023107678}, while the study by Gnanaolivu et al. (2024) uses AlphaFold2-predicted protein structures and thermodynamic stability tools to evaluate variant-driven loss-of-function in cancer genes \cite{Gnanaolivu2024.06.03.597089}. Finally, by combining RAG from variant annotations from various sources and fine-tuning, GPT-based models have been adapted to annotate and interpret genetic variants \cite{10.1093/bioadv/vbaf019}. Together, these models enable both mechanistic understanding and variant prediction across a range of biological systems.

Transformer-based models are increasingly used to bridge genotype and phenotype by modeling the complex relationships between genetic variation and disease manifestation. Several works focus on polygenic risk score (PRS) prediction: Epi-PRS \cite{Zeng2024.10.04.24314860} and epiBrainLLM \cite{Liu2024.10.03.24314824} leverage genomic large language models to convert personal sequences into intermediate epigenomic signals, improving disease risk prediction in breast cancer, diabetes, and Alzheimer’s disease. Genotype-to-Phenotype Transformer (G2PT) introduces a hierarchical transformer to model gene-function-phenotype pathways and uncover epistatic interactions \cite{Lee2024.10.23.619940}. For direct disease classification, Genomics Transformer models SNP-SNP interactions for Parkinson’s diagnosis \cite{9926815}, while Prophet (Predictor of
phenotypes) predict cellular phenotype (e.g. gene expression, cell viability, and cell morphology) by learning relationships between cellular state, the treatments being performed, and the intended phenotype \cite{Ji2024.08.12.607533}. FREEFORM (Free-flow Reasoning and Ensembling for Enhanced Feature Output and Robust Modeling) \cite{lee2025knowledgedrivenfeatureselectionengineering} applies chain-of-thought prompting to enhance feature selection in genotype-phenotype models, while another research benchmarked a lot of models (TabTransformer \cite{huang2020tabtransformertabulardatamodeling}, FT-Transformer (Feature Tokenizer + Transformer) \cite{gorishniy2023revisitingdeeplearningmodels} and others) for genome-wide prediction in tabular genomic data \cite{Fan2024}. Other studies highlight multimodal and knowledge-enhanced approaches: Shirkavand et al. (2023) propose a multimodal integration to a transformer-GAN model for brain disorder prediction using neuroimaging and genetics \cite{shirkavand2023incompletemultimodallearningcomplex}, and GENEVIC (GENetic data Exploration and Visualization via Intelligent interactive Console) integrates ChatGPT with biomedical knowledge from PubMed, Google Scholar, and arXiv to support variants exploration in complex diseases \cite{10.1093/bioinformatics/btae500}. These models demonstrate the growing role of transformers and LLMs in phenotype prediction, enabling more personalized and mechanistic insights across human genetics.

\paragraph{Clinical Variant Interpretation.}

 The last step of analysis is the interpretation of extracted genetic variants, where the goal is not to discover new variation, but to determine which observed variants are causal or clinically relevant. This section highlights recent work on two main tasks: (1) identifying and classifying disease-causing variants, and (2) prioritizing variants or genes based on phenotypic evidence.

 For variant identification and classification, Genetic Transformer (GeneT) uses a fine-tuned LLM to emulate expert reasoning in the identification of causative variants in rare genetic diseases, leveraging semantic understanding and phenotype-aware context to reduce the variant search space \cite{Liang2024.07.18.24310666}. Unlike traditional filtering pipelines, it learns prioritization patterns directly from data, offering a scalable alternative to manual curation. In parallel, MAVERICK (Mendelian Approach to Variant Effect pRedICtion built in Keras) uses transformer ensembles to assess and classify the pathogenicity of variants in Mendelian (monogenic) disorders \cite{Danzi2023}. Boulaimen et al. (2024) integrate DNA, protein, and structural representations through multiple state-of-the-art transformer-based models to classify variants, particularly those that are with uncertain significance, by capturing cross-modal dependencies \cite{boulaimen2024integratinglargelanguagemodels}. Another notable tool, VarChat, enables a user-friendly interaction with the generative AI model in natural language, automating the summarization of variant-related scientific literature, helping users contextualize variant effects with up-to-date knowledge from trusted publications \cite{De_Paoli2024-jy}.

Another instrument, PhenoSV, applies attention-based modeling to structural variants (SVs) using transformer mechanisms to capture how both non-coding and coding structural variants affect gene function \cite{Xu2023-tq}. MGI (Multimodal contrastive learning of Genomic and medical Imaging) fuses medical images and gene expression data using attention mechanisms, enabling downstream gene prioritization tasks \cite{zhou2024mgimultimodalcontrastivepretraining}. Kim et al. (2024) evaluate general-purpose LLMs (including GPT-4 and LLaMA2 (Large Language Model Meta AI) \cite{touvron2023llama2openfoundation}) for gene prioritization based on phenotype descriptions (provided as both structured inputs (e.g., HPO terms) and free-text clinical narratives), highlighting their high risk of bias and limited ability to generalize \cite{Kim2024}. Across these efforts, transformers offer a powerful framework to connect variants, genes, and phenotypes—capturing context, structure, and relationships that are difficult to encode with traditional rules or statistical filters.

\subsubsection{Post-Analytical stage: diagnosis and decisions}

This section describes the post-analytical phase, which encompasses the synthesis and contextualization of diagnostic outputs into clinically actionable insights. At this stage, transformer-based models are applied to tasks that extend beyond pure prediction: allowing disease subtyping, patient stratification, integration of multimodal data for prognosis and treatment recommendations, and the generation of clinical reports. These applications emphasize the interpretability, generalizability, and semantic reasoning capabilities of transformer models, offering a new layer of diagnostic intelligence that aligns computational predictions with established medical knowledge and decision-making frameworks.

\paragraph{Patient Clustering and Subtyping.}

Transformers have emerged as powerful tools for stratifying patients and identifying disease subtypes by integrating imaging, textual, and longitudinal clinical data. In oncology, Wang et al. (2024) proposed ETMIL-SSLViT (Ensemble Transformer-based Multiple Instance Learning with Self-Supervised Learning Vision Transformer feature encoder), a framework that predicts both pathological subtypes and tumor mutational burden (TMB) in endometrial and colorectal cancer patients directly from H\&E-stained whole-slide images. This image-only approach bypasses the need for sequencing while aiding effective immunotherapy planning \cite{WANG2025103372}. In the neurodegeneration domain, Tri-COAT (tri-modal co-attention mechanism) is a multimodal transformer that subtypes Alzheimer's disease patients using early-stage imaging, genetic, and clinical data. They capture biologically grounded cross-modal interactions and identify slow, intermediate, and fast-progressing subtypes \cite{jpm14040421}. Another approach employs LLMs integrated with RAG to classify central nervous system tumours from free text histopathological reports, using current WHO guidelines \cite{https://doi.org/10.1002/2056-4538.70009}. Finally, Qiu et al. (2024) developed TransVarSur, a transformer-variational encoder tailored for longitudinal survival data from EHRs. Applied to Crohn’s disease, the model identified clinically meaningful patient subgroups with divergent survival trajectories, bridging clustering with downstream event prediction \cite{Qiu2024.01.11.24301148}.

\paragraph{Data and Results Aggregation.}

In the post-analytical phase of clinical genomics and diagnostics, a key challenge is the integration of heterogeneous data sources, ranging from sequencing outputs to clinical histories, into meaningful predictions or recommendations, including extracting interpretation-relevant knowledge, predicting patient trajectories and clinical outcomes, and proposing evidence-based treatment strategies. 

Stroganov et al. (2023) applied LLM to structure diagnostic outputs, more specifically, neuropathological reports of Parkinson’s disease patients from the NeuroBioBank were used to extract anatomical and pathological findings (both microscopic and macroscopic) into a standardized, interoperable format aligned with Common Data Elements \cite{Stroganov2023.09.12.557252}. Complementing this, Wang et al. (2024) proposed a question-answering pipeline using ChatGPT to extract pathogenic microorganism knowledge from mNGS literature. While not directly related to human genetics, their workflow (based on a fine-tuned Q\&A model and data augmentation) offers a compelling demonstration of automated interpretation techniques that could be adapted to human variant interpretation pipelines \cite{Wang2024-hv}.

A growing body of work applies transformers to predict health trajectories and clinical outcomes from large-scale EHR data. Zhou et al. (2024) introduced CPAE, a Contrastive Predictive AutoEncoder designed for unsupervised pre-training on EHRs, which predicts in-hospital mortality and length-of-stay even with limited labeled data \cite{zhou2024mgimultimodalcontrastivepretraining}. Yang et al. (2024) developed ForeSITE (Forecasting Susceptibility to Illness with Transformer Embeddings), system to model multimorbidity trajectories in the UK Biobank by capturing co-occurrence patterns of chronic diseases over time \cite{Yang2024.10.02.24314786}. Manzini et al. (2025) presented DARE (Diabetic Attention with Relative position Representation Encoder), which is pre-trained on longitudinal diabetes data and fine-tuned for clinical outcome prediction (comorbidity occurrence and glycemic control) \cite{MANZINI2025126876}. 

The application of generative transformer models for personalized treatment recommendation is a nascent but promising domain. Hamilton et al. (2024) \cite{Hamilton2024-cw} conducted a comparative evaluation of ChatGPT models on next-generation sequencing reports in oncogene-driven non-small cell lung cancer. Using a novel Generative AI Performance Score (G-PS), the study found GPT-4 to produce more accurate and guideline-compliant treatment suggestions than GPT-3.5, with fewer hallucinations, exemplifying how generative LLMs can be systematically benchmarked for clinical relevance  \cite{Hamilton2024-cw}.

\paragraph{Clinical Report Generation and Decision Support.}

In the final stage of the diagnostic pipeline, transformer-based systems are applied to support clinical decision-making by automating report generation, synthesizing genomic data, and enabling communication with patients and professionals. 

An early example of report-oriented tooling is Just-DNA-Seq, an open-source platform designed to generate user-friendly genomic reports from personal VCF files, with a focus on longevity-related annotations. In addition to variant annotation and polygenic risk scoring, the platform includes a GPT-based module (GeneticsGenie) for answering user queries based on underlying interpretation modules, showcasing how LLMs can be paired with rule-based genomic annotation engines to support direct-to-consumer interpretation workflows \cite{anton2024justdnaseqopensourcepersonalgenomics}. A related effort by Mastrianni et al. (2024) investigates how clinical and research experts in genetics expect to use generative AI tools in their professional workflow. The authors identified use cases (supporting information gathering, interpretation, and decision-making across stages of the analytics pipeline) and concerns (including trust, transparency, integration into existing systems) \cite{mastrianni2024aienhancedsensemakingexploringdesign}. Additionally, Lantz et al. (2023) demonstrated the use of ChatGPT in co-authoring a case report on toxic epidermal necrolysis, showcasing its utility in drafting clinical narratives while underscoring the necessity of expert oversight due to occasional factual inaccuracies \cite{Lantz2023-ub}.

Several studies evaluate how LLMs might operate as communicative agents: either assisting doctors or directly interacting with patients. In oncology, Lukac et al. (2023) assessed ChatGPT’s ability to contribute to multidisciplinary tumor board decisions for breast cancer patients: it lacked specificity and occasionally misclassified biomarker interpretations, highlighting current limitations in reliability for case-level recommendations \cite{Lukac2023}. In contrast, Quidwai et al. (2024) developed a RAG chatbot tailored to multiple myeloma, integrating Mistral-7B \cite{jiang2023mistral7b} with domain-specific knowledge graphs and embedding models to offer personalized treatment plans based on patient-specific data \cite{Quidwai2024.03.14.24304293}. In contrast, several works benchmark LLMs' decision-making capabilities across tasks: GPT-4 on NEJM (the New England Journal of Medicine) quiz-style diagnosis questions \cite{Ueda2024}, or different GPT models on phenotypic case reports \cite{Reese2024.07.22.24310816}, observing different level of accuracy (from near-clinical-level to low) in some cases. In the context of pharmacogenomics, Keat et al. (2025) introduced PGxQA (PharmacoGenetiCS Question-Answering), a benchmark evaluating how LLMs respond to stakeholder-specific pharmacogenomic questions, revealing persistent challenges in alignment with medical consensus \cite{Keat2025-lv}.

Several studies also explore LLMs in decision support for precision oncology. A multi-model comparison across ChatGPT, Galactica \cite{GALACTICA}, Perplexity \cite{perplexity2025}, and BioMedLM \cite{bolton2024biomedlm27bparameterlanguage} showed that while LLMs proposed more treatment options than human experts, they often lacked supporting evidence or clinical rigor (however, some of them are considered helpful by tumor board members) \cite{10.1001/jamanetworkopen.2023.43689}. Fukushima et al. (2025) evaluated domain adaptation strategies for a Japanese LLM aimed at supporting genetic counseling. Different tuning methods were tested, RAG offered the best balance between accuracy and medical appropriateness according to this research \cite{Fukushima2025-oi}.  A complementary benchmarking effort by Patel et al. (2024) showed that ChatGPT-3.5 (currently outdated) has a high ability to answer genetic counseling questions for gynecologic cancers both correctly and comprehensively \cite{PATEL2024115}. Meanwhile, Luca et al. (2023) conducted qualitative interviews with patients and families undergoing genetic testing, researching whether chatbots are acceptable by patients: they favored chatbots as adjuncts, not replacements, especially if they knew they could fall back on or consult a real clinician if needed \cite{Luca2023-st}.

\subsection{Education}

The use of LLMs and generative AI in education, especially in the field of genetics, offers significant potential. These models help in teaching both medical students and professionals. $9$ articles in this systematic review highlight three major directions in which LLMs and generative AI contribute to education: conversational and Q\&A learning systems, generation of synthetic visual materials, support for collaborative learning and group formation. 

Dialogue-based learning with LLMs is emerging as a practical alternative to traditional study partners or tutors, especially for genetics and rare diseases. A range of studies have evaluated ChatGPT’s performance in answering structured questions across internal medicine, genomics, and patient education. While the last GPT models of OpenAI show clear gains over earlier versions, studies reveal variability in accuracy, especially in nuanced topics, such as inheritance patterns or ethical subtleties of genetic risk communication \cite{McGrath2024-zx, Walton2023.10.25.564074, Hernandez2023-ay}. A key trend is the use of retrieval-augmented generation (RAG) and prompt engineering to enhance domain specificity, suggesting that medical knowledge augmentation can significantly improve educational performance \cite{Tarabanis2024-er, weng2023largelanguagemodelsneed} - these techniques will be considered in the discussion more in-depth. Despite these advances, models still risk hallucination and outdated references, highlighting the need for oversight and continual retraining \cite{Walton2023.10.25.564074, Hernandez2023-ay}. The utility of LLMs lies not in full replacement of educators but in providing on-demand, interactive guidance that reinforces core concepts and helps bridge gaps in access to expert instruction.

Generative AI offers a promising solution to the ethical and privacy challenges of using real patient images in genetics education. A study on Kabuki and Noonan syndromes found that AI-generated facial images, created using StyleGAN \cite{karras2019stylebasedgeneratorarchitecturegenerative} methods, were nearly as effective as real photos in training pediatric residents to recognize phenotypic features \cite{Waikel2024-na}. While real images were rated slightly more helpful, synthetic ones notably increased diagnostic confidence and reduced uncertainty. Additionally, another research highlights the possibility of patient information leaflets (PILs) generation using ChatGPT \cite{Verran2024-en}. These findings highlight the potential of the use of generated images as a privacy-safe, scalable adjunct in medical training, especially for rare or underrepresented conditions.

Together, these approaches offer scalable, privacy-preserving, and adaptive tools to complement existing curriculum and address the shortage of specialized genetics educators.

\section{Discussion}

As transformer models and, more specifically, LLMs (e.g. GPT, BERT) are increasingly integrated into specialized fields such as genetics, it is essential to evaluate their strengths and limitations. This research systematically reviewed the application of LLMs and generative AI models in genetic diagnostics, incorporating resources from PubMed, bioRxiv, medRxiv, and arXiv to ensure both peer-reviewed depth and inclusion of the latest model developments.

In this section, we outline additional areas of genetics where generative AI show strong potential but were beyond the scope of this review. We also highlight key developments needed to ensure the reliability and trustworthiness of LLM applications, and discuss emerging trends and techniques that may enhance their effectiveness.

\subsection{LLM applications in adjacent fields of research}

Although this review focuses on the applications of generative AI models in human genetics and diagnostics, several adjacent research areas, while not directly related to human genome analysis, offer valuable insights and transferable lessons. These applications were excluded from the main focus due to limited direct relevance; however, they highlight the variety of possible usage across biological and medical domains and may inspire future applications in human genetic research.

Studies applying LLMs to microbial genomes have demonstrated the potential of language models to encode meaningful representations of whole genomes. For example, models trained on bacterial or fungal species can predict traits such as antibiotic resistance or habitat specificity \cite{Naidenov2024.03.18.585642, li2025genometransformergeneinteraction, Weinstock2024.12.12.628183}. While distinct from human genetics, these works show how transformer-based models can capture population structure and gene interactions in complex biological systems.

Transformer models have also been applied to protein sequences for predicting gene ontology terms and functional annotations \cite{tamir2024protgotransformerbasedfusion}. These studies operate in the space of proteomics, yet demonstrate modeling principles that could be extended to human gene function prediction or variant interpretation.

A substantial body of work with ViTs focuses on cancer imaging, particularly for tasks such as tumor segmentation, subtype classification, and spatial analysis from whole-slide images \cite{Lo2024-rh, Li2023-zk, Pizurica2024, Hu2024-qo, 10.3389/fninf.2024.1444650, jpm14101022, YANG2024108400}. While not always grounded in genomic data, these tasks intersect with genetic diagnostics when molecular subtypes, such as microsatellite instability, play a role in treatment stratification.
  
Epigenetic regulation and cross-species prediction of gene expression using sequential and imaging data represent another promising direction \cite{Weinstock2024.12.12.628183, RAMPRASAD2024100347, Pizurica2024}. These studies explore how attention-based models can generalize across evolutionary distances, enabling predictions in under-characterized organisms and informing functional annotation pipelines.

LLMs are being increasingly integrated into gene editing workflows: from automating guide RNA design and protocol generation (e.g., CRISPR-GPT) to predicting cellular responses to perturbations at single-cell resolution (e.g., scLAMBDA) \cite{huang2024crisprgptllmagentautomated, Wang2024.12.04.626878}. Similarly, transformers have advanced splice site prediction for identifying disease-relevant splice variants \cite{Jonsson2024}. Together, these applications illustrate how transformer models support both the interpretation and manipulation of gene function in human genetics.

Transformer models have also been used to investigate the role of genetic support in the success or failure of clinical trials \cite{Razuvayevskaya2024}. While not directly diagnostic, such applications emphasize the growing role of human genetic evidence in pharmaceutical development and clinical decision-making.

Taken together, these diverse research directions extend the scope of transformer-based models beyond traditional genetics. By leveraging techniques and datasets from related fields, such as microbial biology, cancer diagnostics, and synthetic biology, future work in human genetics may benefit from models and insights developed in adjacent domains.

\subsection{Data and Benchmarks}

The growing use of generative AI is closely related with the quality of available datasets and benchmarks. Reliable evaluation and generalization critically depend not only on model design but also on data diversity, integrity, and task-relevant benchmarking protocols.
 
LLMs' applications in genetic diagnostics requires reliability, therefore robust benchmarks are vital for comparing models and ensuring reliability. CARDBiomedBench \cite{Bianchi2025.01.15.633272} exemplifies this shift, offering a multi-domain Q\&A benchmark in biomedicine. Its design is based on curated expert knowledge and data augmentation, which exposes real gaps in model reasoning and safety, even among state-of-the-art systems. Amount of benchmarks, scores and proposed ways to track the models' quality rise up \cite{Labbe2023-lj, Tarabanis2024-er, Keat2025-lv, Hamilton2024-cw, Li2024.10.29.621006, murphy2024.06.10.24308475, osullivan2024democratizationsubspecialitymedicalexpertise}. Such domain-specific evaluations help move beyond general-purpose NLP benchmarks and focus attention on the nuanced reasoning required in biomedical decision-making.

Recent work in other technical domains has highlighted the threat of benchmark leakage, where models inadvertently see test data during pretraining \cite{zhou2025lessleakbenchinvestigationdataleakage, ni2025trainingbenchmarkneed}. These studies demonstrate that such leakage can inflate model performance and undermine benchmark credibility, proposing solutions to measure real metrics. They also stress the need for transparent documentation of training data and rigorous, leakage-aware evaluation protocols, especially in sensitive domains, such as biomedicine.

We mentioned multiple times, that modern clinical models must integrate diverse data types: text, images, genomics, structured records, which requires both scalable architectures and consistent input quality. The new methods are being proposed to improve efficiency when fusing many modalities, such as contrastive learning and other methods \cite{golovanevsky2024oneversusothersattentionscalablemultimodal, jpm14040421, wu2024gestaltmmlenhancingraregenetic, Mondol_2024, zhou2024mgimultimodalcontrastivepretraining, shirkavand2023incompletemultimodallearningcomplex, mondol2024mmsurvnetdeeplearningbasedsurvival, Machado_Reyes2024.11.02.24316653}. Additionally, some methods propose preprocessing steps for specific types of data (e.g. segmentation \cite{Yuan2024-wp, shi2023nextouefficienttopologyawareunet, FU2024107938}, or facial axes standardization  \cite{alomar2024automaticfacialaxesstandardization}).

Together, these developments underscore that the value of LLMs in genetics is not solely defined by model architecture. Equally important are the integrity of training and evaluation datasets, the representativeness of benchmarks, and the methods used to integrate and align multimodal inputs. 

\subsection{Biases}

Despite their impressive capabilities, large language models often reflect biases present in their training data, which can affect clinical utility. Several studies have revealed racial and demographic biases in generated medical reports and other types of data \cite{Yang2024-zv, Lin2024-yg}, while others show variations in performance across age-specific manifestations of genetic disorders \cite{Othman2025.01.19.25320798} or reviewer experience levels \cite{LEVIN2024669}. Language also remains a critical source of disparity: most biomedical models are English-centric, limiting accessibility and accuracy in other languages. Resources, such as MedLexSp for Spanish \cite{Campillos-Llanos2023}, Chinese medical conversational Q\&A \cite{weng2023largelanguagemodelsneed}, and domain adaptation efforts for Japanese genetic counseling \cite{Fukushima2025-oi} demonstrate how localized models and lexicons can help reduce these gaps. Overall findings underscore the need for language-specific resources, ongoing fairness audits, and rigorous ethical evaluation.

\subsection{Model Strategies}

Lastly, we discuss methods to assess the effectiveness of transformers models in genetics and related fields, exploring best practices for maximizing the benefits of LLMs while addressing their shortcomings.

Original papers often compared GPT-3.5 and GPT-4, which now are outdated: newer models that combine reasoning and other features have been published. Despite these models have shown good performance in some tasks, studies consistently highlight issues such as hallucinations, outdated information, and stylistic artifacts that make their outputs distinguishable from human experts \cite{PATEL2024115, Hulman2023-zp, McGrath2024-zx, hier2024highthroughputphenotypingclinicaltext}. Using the latest model versions, fine-tuning them on domain-specific data, and thorough prompting improves performance, but careful evaluation remains necessary \cite{Mondillo2024.08.20.24312291, https://doi.org/10.1002/2056-4538.70009, genes16010029, Temsah2024-ux}.

Choosing the right architecture is critical. BERT-based models excel at information extraction tasks such as named entity recognition or relation extraction, making them suitable for structuring clinical data pr texts. In contrast, GPT models are better for generative tasks like summarization or dialogue, while full encoder-decoder architectures may be preferable for interactive tools or personalized report generation. Some studies simply use decoder-only models (e.g. chatGPT) for their research, although using a transformer encoder would possibly yield better results \cite{Labbe2023-lj, Shringarpure2024.05.30.24308179, hier2024highthroughputphenotypingclinicaltext}.

Prompting significantly affects output quality. RAG-based systems that retrieve real-time evidence before generating responses could improve factual consistency \cite{Coen2024-ug, Murugan2024-lo, Fukushima2025-oi, Quidwai2024.03.14.24304293, https://doi.org/10.1002/2056-4538.70009, 10.1093/bioadv/vbaf019}. Expert-reviewed prompt engineering pipelines have already been deployed in use cases, such as returning positive genetic screening results \cite{Coen2024-ug}, medical Q\&A \cite{Fukushima2025-oi}. These approaches help mitigate the risk of relying solely on a model’s internal memory.

Emerging advanced techniques, such as chain-of-thought (CoT) prompting and LLM-based agents further enhance performance on complex tasks. Models equipped with tool-use capabilities (executing code, querying biomedical databases, or self-criticizing their outputs) show promise in hypothesis generation and experiment planning. Notable examples include BioDiscoveryAgent for genetic perturbation experiments design \cite{roohani2025biodiscoveryagentaiagentdesigning} or a chatbot agent for facilitating family communication of hereditary risk in familial hypercholesterolemia \cite{WALTERS2023100134}. These agents illustrate how LLMs can be embedded into structured decision-support workflows.

Recent findings suggest that domain-specific pretraining alone does not guarantee superior performance: randomly initialized models can match or exceed genomic foundation models in downstream tasks \cite{Vishniakov2024.12.18.628606}. Instead, training strategy plays a critical role. Two primary methods are mixed training and train-and-finetune strategies. Mixed training exposes the model to a diverse range of data and tasks simultaneously, enabling the development of more generalized representations. This approach helps models avoid overfitting while enhancing memorization and flexibility. In contrast, the traditional train-and-finetune method focuses on a specific task during fine-tuning, which can lead to deeper, more domain-specific knowledge but carries a risk of overfitting. Allen-Zhu’s research emphasizes that mixed training produces models that retain broader, flexible knowledge for downstream tasks such as retrieval and classification, making them more robust for real-world applications where adaptability is key \cite{allenzhu2024physicslanguagemodels32}. This balance between specialization and generalization is crucial for domains like genetics, where the knowledge base is continually evolving, and maintaining up-to-date, accurate information is essential for applications such as genetic counseling or variant interpretation.

Together, these insights highlight that the effectiveness of LLMs and generative AI in genetics depends not just on model scale, but on task alignment, prompt structure, real-time access to knowledge, and interactive reasoning tools, all of which are key to trustworthy clinical deployment.

\section{Conclusions}
As detailed in this review, transformer-based models have made significant progress in various critical tasks within the research and diagnosis of human genetic diseases. They have excelled in genomic variant identification, annotation, and interpretation. LLMs have also greatly improved data extraction from unstructured medical texts, facilitating the organization of clinical and genetic data. Additionally, LLMs are successfully applied in medical imaging through ViTs (e.g. tumor segmentation, facial phenotyping for genetic syndromes). Finally, the integration of LLMs in post-analytical, including patient clustering and decision support, has contributed to more efficient diagnostic workflows, enabling the stratification of patients based on genetic, clinical, and imaging data.

Beyond their integration into the diagnostic pipeline, these tools hold clear potential for supporting various professional roles involved in genetic medicine. For clinical geneticists, LLM-powered systems (described in this article as well as newly developed) can assist in case-specific usage of genetic data and finding evidence from the literature. For laboratory geneticists, transformer models can streamline variant classification, automate reporting, and improve consistency in annotation. For bioinformaticians, such models offer solutions for knowledge extraction, integration of multimodal data, and augmentation of decision-support pipelines. As LLMs mature, we anticipate their deployment in software environments designed to assist these distinct expert groups, enhancing the quality and speed of inherited disease diagnostics.

Naturally, this review cannot cover every tool and model in a field that evolves so rapidly. Rather, it provides a structured overview that can serve as a classifier and guide, helping researchers and practitioners navigate the fast-growing landscape of LLM applications in genetics and diagnostics.

\section{Data Availability}
All data and code pertinent to the results presented in this work are available at \url{https://github.com/TohaRhymes/llm_in_diagnostics}.

\renewcommand{\thefigure}{\arabic{figure}}
\renewcommand{\figurename}{Supplementary Figure}
\setcounter{figure}{0}

\section{Acknowledgments}
This research was supported by the Ministry of Science and Higher Education of the Russian Federation (project “Multicenter research bioresource collection “Reproductive Health of the Family” contract No. 075-15-2025-478 from 29 May 2025).

\newpage
\bibliographystyle{unsrt}  
\bibliography{references}

\section{Appendix}

\subsection{Appendix A: TF-IDF-Based Semantic Phrase Extraction}

\begin{figure}[!ht]
    \centering
    \includegraphics[width=0.6\textwidth]{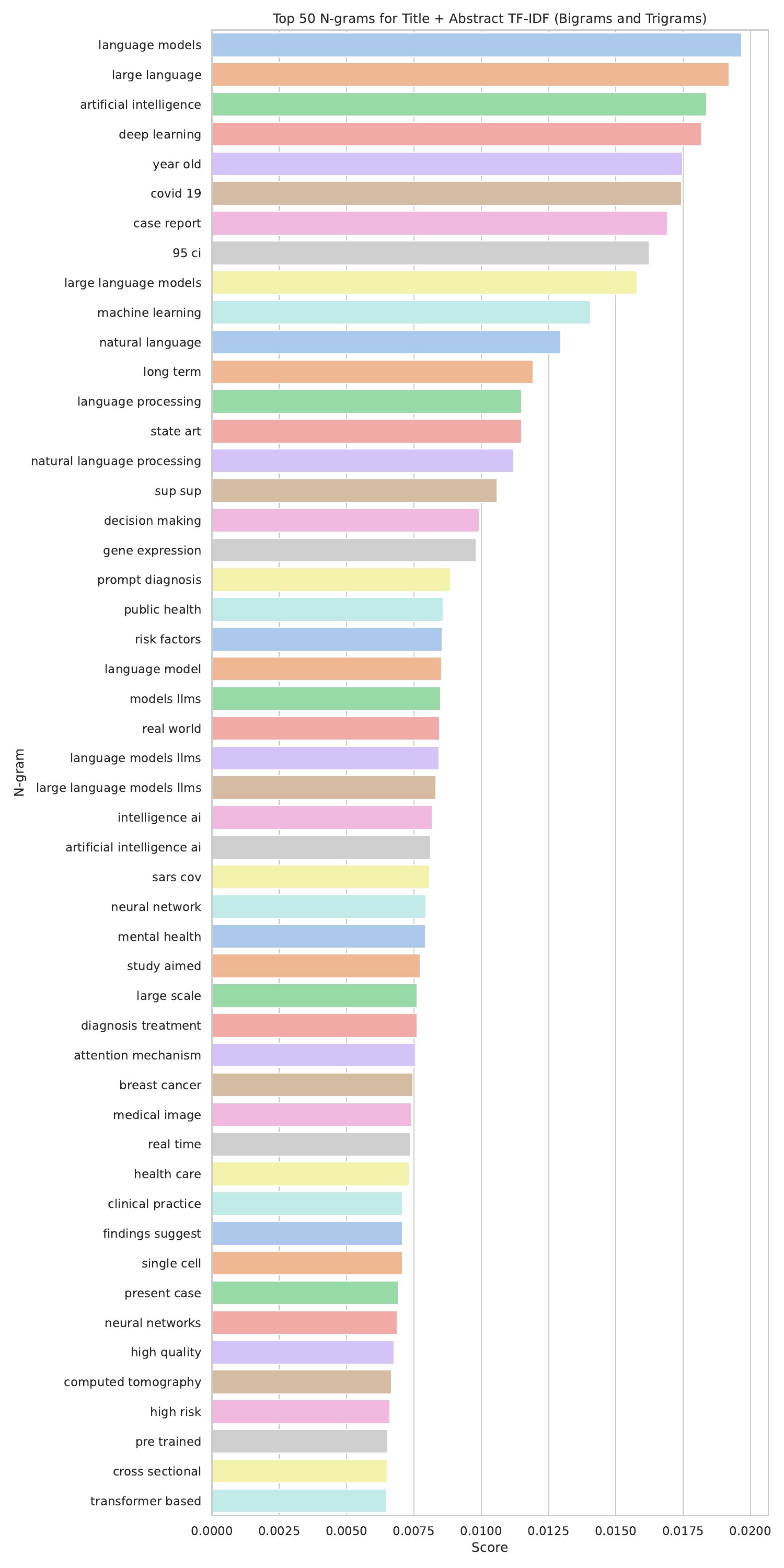}
    \caption{Top 50 TF-IDF-ranked n-grams (bigrams to trigrams). This figure shows the most informative 2–3-word phrases extracted from titles and abstracts using TF-IDF. These terms were used to define thematic groups related to LLMs, genetics, and clinical medicine.}
    \label{SF1}
\end{figure}

\begin{figure}[!ht]
    \centering
    \includegraphics[width=0.6\textwidth]{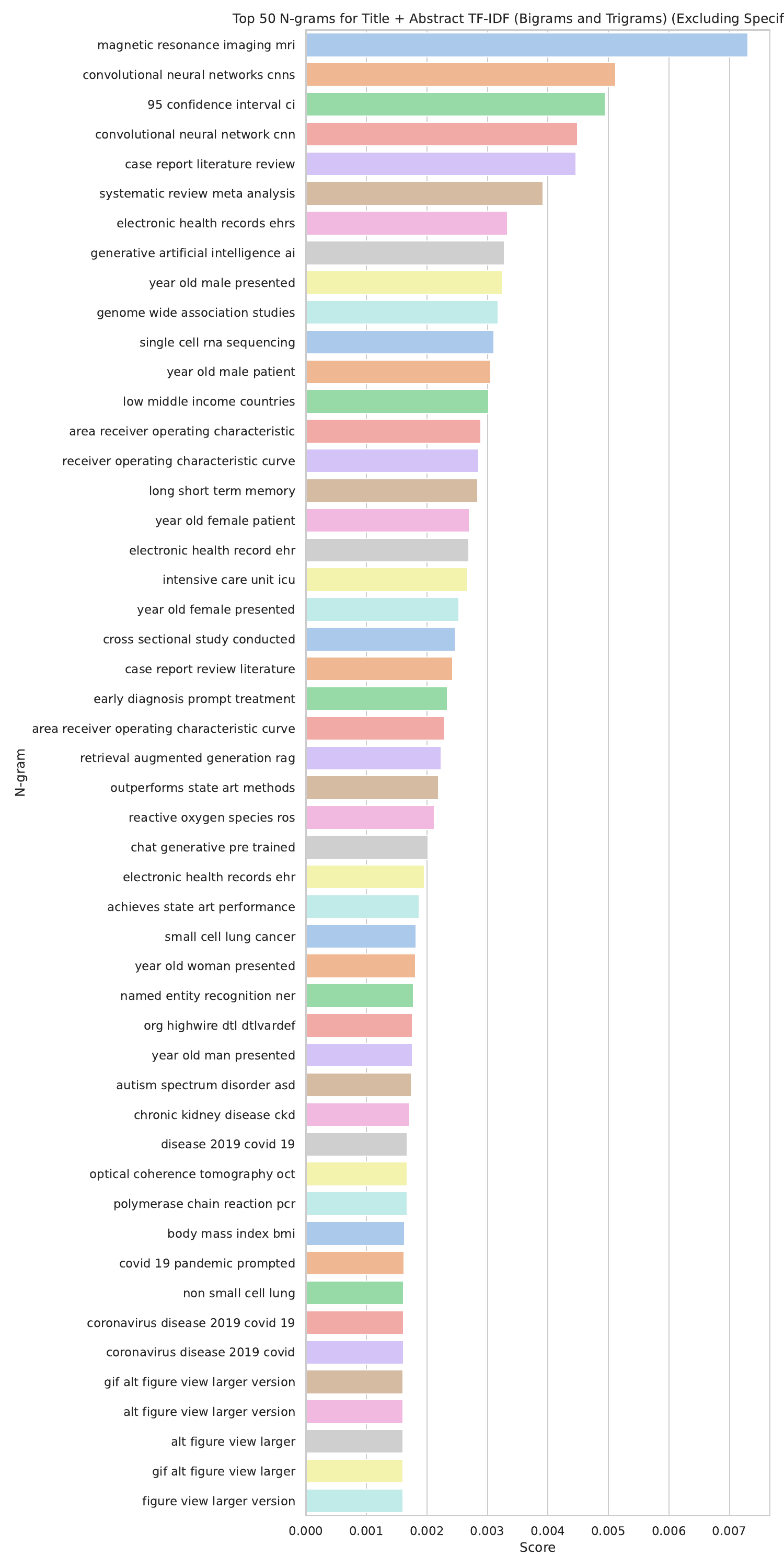}
    \caption{Top 50 TF-IDF-ranked n-grams (4–6-grams), excluding top (common) phrases from Figure \ref{SF1}. This figure shows longer multi-word expressions (4–6 grams) that provide additional context and semantic structure, excluding high-frequency shorter phrases identified earlier.}
    \label{SF2}
\end{figure}

To identify domain-relevant terminology in the article corpus, we applied TF-IDF analysis to combined titles and abstracts, using $n$-gram tokenization. Preprocessing involved HTML tag removal, lowercasing, and merging near-duplicate entries based on title and abstract similarity. Two sets of $n$-grams were extracted: common phrases (2–4 grams) for Supplementary Figure \ref{SF1}, and longer expressions (4–6 grams) for Supplementary Figure \ref{SF2}. In the latter, overly generic terms such as “language”, “model”, and “LLM” were excluded to better emphasize domain-specific terminology. The top 50-ranked phrases in each case were visualized using bar plots.

\subsection{Appendix B: Supplementary Tables – Annotated Article Dataset}

Two supplementary tables were compiled to support the analysis presented in this study. Supplementary Table~1 contains the complete list of articles included after initial collection, deduplication, and manual verification. Supplementary Table~2 provides an extended, manually annotated version of the dataset with additional semantic tags and classification columns.

\textbf{Supplementary Table 1} (ST1) presents the cleaned dataset after the removal of duplicates and initial triage. Duplicate entries were identified not only through automatic preprocessing but also through joint manual assessment by two researchers, ensuring a consistent and conservative approach to inclusion. ST1 includes metadata such as the article title, abstract, source, review status, and initial relevance tag.

\textbf{Supplementary Table 2} (ST2) expands upon this initial dataset by including additional annotations used in the systematic analysis. These include section-level classification (\texttt{what section used}), fine-grained labels for specific tasks inside these stages (\texttt{subgroup}), topic tags related to artificial intelligence and medicine (\texttt{ai\_topic}, \texttt{medicine\_topic}), and three binary relevance flags (\texttt{not\_relevant}, \texttt{partly\_relevant}, \texttt{relevant}). Five manually selected articles were also added at this stage (four highly relevant and one partially relevant), resulting in a total of 304 articles in ST2. These additions were motivated by expert review and targeted searches within the originally collected corpus and cited references. 

Detailed descriptions of column meanings and classification codes are available in the project GitHub repository\footnote{\url{https://github.com/TohaRhymes/llm_in_diagnostics}}.









\end{document}